\documentclass{article}

\usepackage{microtype}
\usepackage{graphicx}
\usepackage{booktabs} 
\usepackage{caption}
\usepackage{enumitem}
\usepackage{multirow}       
\usepackage{makecell}       
\usepackage{xcolor}
\usepackage{subcaption}
\usepackage{url}
\usepackage{listings}
\usepackage{multirow}
\usepackage{etoolbox}
\newcommand{\zerodisplayskips}{%
  \setlength{\abovedisplayskip}{0pt}%
  \setlength{\belowdisplayskip}{0pt}%
  \setlength{\abovedisplayshortskip}{0pt}%
  \setlength{\belowdisplayshortskip}{0pt}}
\appto{\normalsize}{\zerodisplayskips}
\appto{\small}{\zerodisplayskips}
\appto{\footnotesize}{\zerodisplayskips}
\usepackage{amsmath,amsfonts,bm}

\def\gL{{\mathcal{L}}}

\def\gS{{\mathcal{S}}}

\def\sD{{\mathbb{D}}}

\def\0{\mathbf{0}}
\def\1{\mathbf{1}}

\usepackage{hyperref}

 \usepackage[accepted]{icml2021}

\icmltitlerunning{Rethinking Co-design of Neural Architectures and Hardware Accelerators}

\begin{document}

\twocolumn[
\icmltitle{Rethinking Co-design of Neural Architectures and Hardware Accelerators}



\icmlsetsymbol{equal}{*}

\begin{icmlauthorlist}
\icmlauthor{Yanqi Zhou}{equal,google} 
\icmlauthor{Xuanyi Dong}{equal,google}
\icmlauthor{Berkin Akin}{equal,google}
\icmlauthor{Mingxing Tan}{google}
\icmlauthor{Daiyi Peng}{google}
\icmlauthor{Tianjian Meng}{google}
\icmlauthor{Amir Yazdanbakhsh}{google}
\icmlauthor{Da Huang}{google}
\icmlauthor{Ravi Narayanaswami}{google}
\icmlauthor{James Laudon}{google}
\end{icmlauthorlist}

\icmlaffiliation{google}{Google, Mountain View, USA}

\icmlcorrespondingauthor{Yanqi Zhou}{yanqiz@google.com}
\icmlcorrespondingauthor{James Laudon}{jlaudon@google.com}

\icmlkeywords{Machine Learning, ICML}

\vskip 0.3in
]




\printAffiliationsAndNotice{\icmlEqualContribution} 

\begin{abstract}
Neural architectures and hardware accelerators have been two driving forces for the progress in deep learning. Previous works typically attempt to optimize hardware given a fixed model architecture or model architecture given fixed hardware. And the dominant hardware architecture explored in this prior work is FPGAs. 
In our work, we target the optimization of hardware and software configurations on an industry-standard edge accelerator. 
We systematically study the importance and strategies of co-designing neural architectures and hardware accelerators. We make three observations: 1) the software search space has to be customized to fully leverage the targeted hardware architecture, 2) the search for the model architecture and hardware architecture should be done jointly to achieve the best of both worlds, and 3) different use cases lead to very different search outcomes.
Our experiments show that the joint search method consistently outperforms previous platform-aware neural architecture search, manually crafted models, and the state-of-the-art EfficientNet on all latency targets by around 1\% on ImageNet top-1 accuracy. Our method can reduce energy consumption of an edge accelerator by up to 2x under the same accuracy constraint, when co-adapting the model architecture and hardware accelerator configurations. 
\end{abstract}

\section{Introduction}
\label{sec:intro}

\begin{figure}[hbt!]
  \centering
      \includegraphics[width=0.95\linewidth,trim={0cm 0cm 0cm, 0cm},clip]{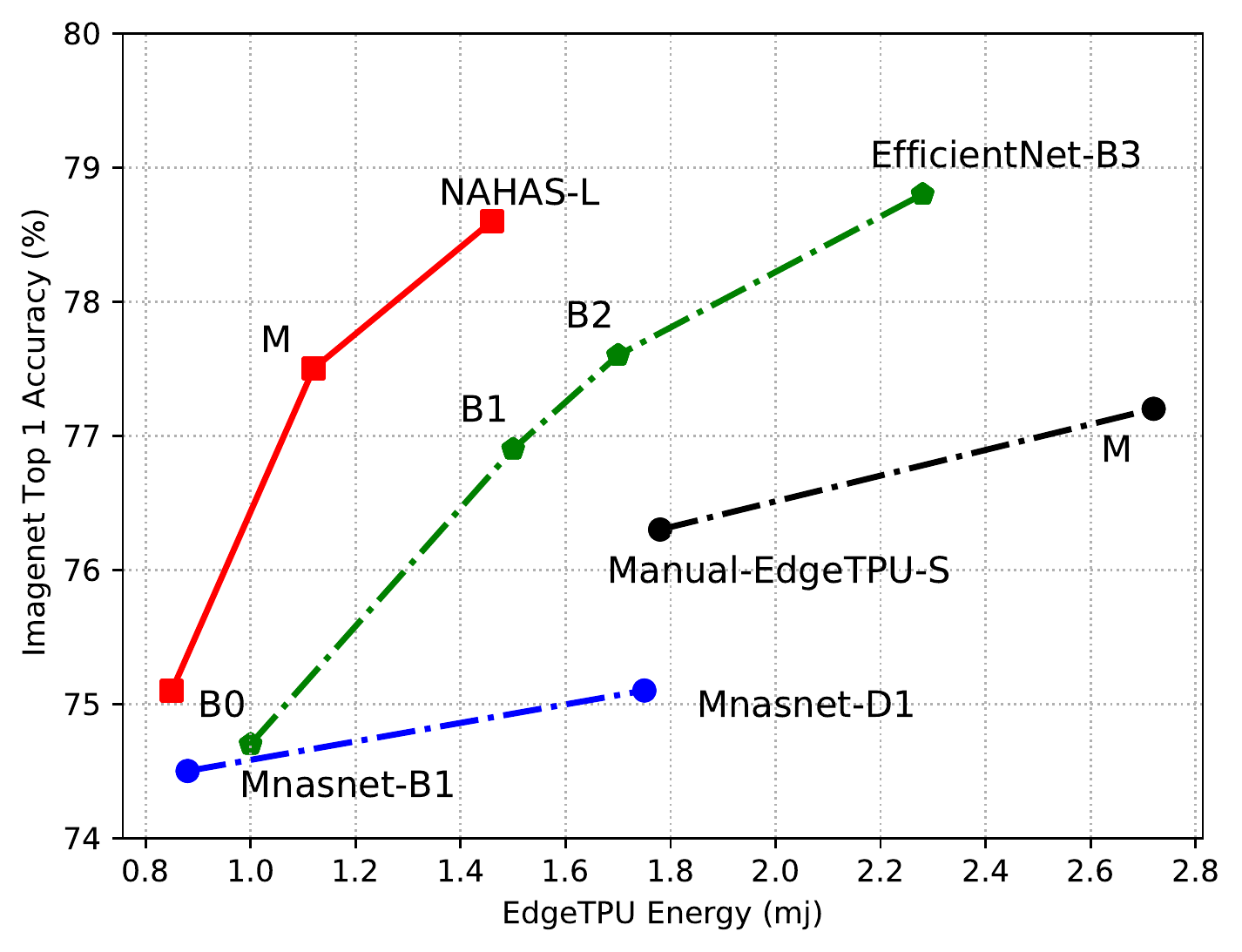}\vspace{-1em}
 \caption{Chip Energy ($Power \times Latency$) vs. ImageNet top-1 accuracy.
 Joint search for hardware and model architectures (NAHAS) significantly outperforms other platform-aware NAS, or manually crafted efficient ConvNets. Manual-EdgeTPU are manually crafted models on an evolved EdgeTPU search space defined in Section~\ref{subsec:evolved_ss}.}
    \label{fig:first_page_results}
\end{figure}

The traditional approach to hardware design is to build a general purpose processor that is optimized for a particular set of workloads. For example, CPUs are optimized for sequential workloads such as desktop applications, and GPUs are designed for massive parallel workloads such as gaming, graphics rendering, scientific computing, etc. The optimization of these designs is typically done by evaluating the performance of the designs on a set of benchmarks (e.g., SPEC~\cite{SPEC2006}), which act as a proxy for the full set of workloads. In addition to the difficulty for the benchmarks to capture all the salient characteristics of the workload set, the benchmarks tend to stay fixed for a substantial period of time, which can result in the hardware design lagging behind the algorithmic changes, particularly in a rapidly changing field such as machine learning. 

As a result of hitting the end of Moore’s Law in the recent decade, the focus for continued processing improvements has moved to hardware specialization to provide additional speedups and efficiency for a narrowed-down application or domain.
Google's TPU~\citep{TPU}, Intel’s Nervana NNP~\citep{IntelNNP}, and Apple's Neural Engine in the M1 SoC~\citep{appleM1} are representative accelerators specialized for deep learning primitives. MLPerf~\citep{MLPerf} has become prevalent for benchmarking the state-of-the-art design of machine learning (ML) accelerators. However, rapid progress in deep learning has given birth to numerous more expressive and efficient models in a short time, which results in both benchmarking and accelerator development lagging behind.
For example, squeeze-and-excite with global pooling and SiLU/Swish non-linearity~\citep{swish2017,swishsil18} are found to be useful in EfficientNet~\citep{EfficientNet2019}; however,  
they are often not supported or inefficient in many specialized accelerators like GPUs, EdgeTPUs and DSPs.

Platform-aware neural architecture search (NAS)~\citep{MnasNet2018,FBNet2019,proxylessNAS2019} optimizes the neural architectures for a target inference device. The target device has a fixed hardware configuration that can significantly limit NAS performance and efficiency. For example, the target device may have a sub-optimal compute and memory ratio for the target application combined with an inference latency target, which can shift the optimal NAS model distributions and result in underperformance.

There is also a shift towards optimizing hardware architecture given a fixed model architecture. For example, design space exploration in computer systems has become more crucial due to the surge of specialized hardware~\citep{bo:frontiers:2020,flexibo:arxiv:2020,cnn_gen:cyber:2020,prac_dse:mascots:2019,accel_gen:dac:2018,spatial:pldi:2018,automomml:hpc:2016,opentuner:pact:2014}. Combining this success with platform-aware NAS, some recent attempts have been made in jointly co-optimizing hardware and model architectures~\citep{jiang2020standing,choi2020dance,jiang2020hardware,achararit2020apnas,coexploration2020,kwon2018co,yang2019synetgy}. However, most of the work target FPGAs or academic accelerators~\cite{Eyeriss2016}, which have some limited capabilities such as not being able to run arbitrary networks on large datasets like ImageNet. They typically study the co-optimization of a fixed neural network without changing the NAS search space.

Building on this prior work, we propose to study the joint search of hardware and model architectures. We generalize the above ideas into a framework called NAHAS -- Neural Hardware and Hardware Accelerator Search. In NAHAS, we parameterize neural architecture search and hardware accelerator search in a unified joint search space.
We use a highly parameterized industry-standard edge accelerator as our target device, which has a tunable set of important hardware configurations.
These knobs fundamentally determine hardware characteristics such as the number of computing units, compute to memory ratio, bandwidth, etc., which we find very critical to model performance.
We formulate the optimization problem as a bi-level optimization with hardware resource constraints on the chip area and model latency or energy (which encapsulates both power and latency).

Unlike the conventional search approach, NAHAS is \textit{task driven}, where the task is a problem (e.g., image classification, object detection) or a domain of problems (e.g., vision, NLP), not a set of fixed programs or graphs (e.g., ResNet, Transformers). This effectively creates generalization across the vertical stack, making the hardware evolve with the applications. NAHAS can be practically used to design customized accelerators for autonomous driving and mobile SoC (system-on-chip), where a set of highly optimized accelerators are combined into a system.

\begin{figure}[hbt!]
  \centering
    \includegraphics[width=\linewidth]{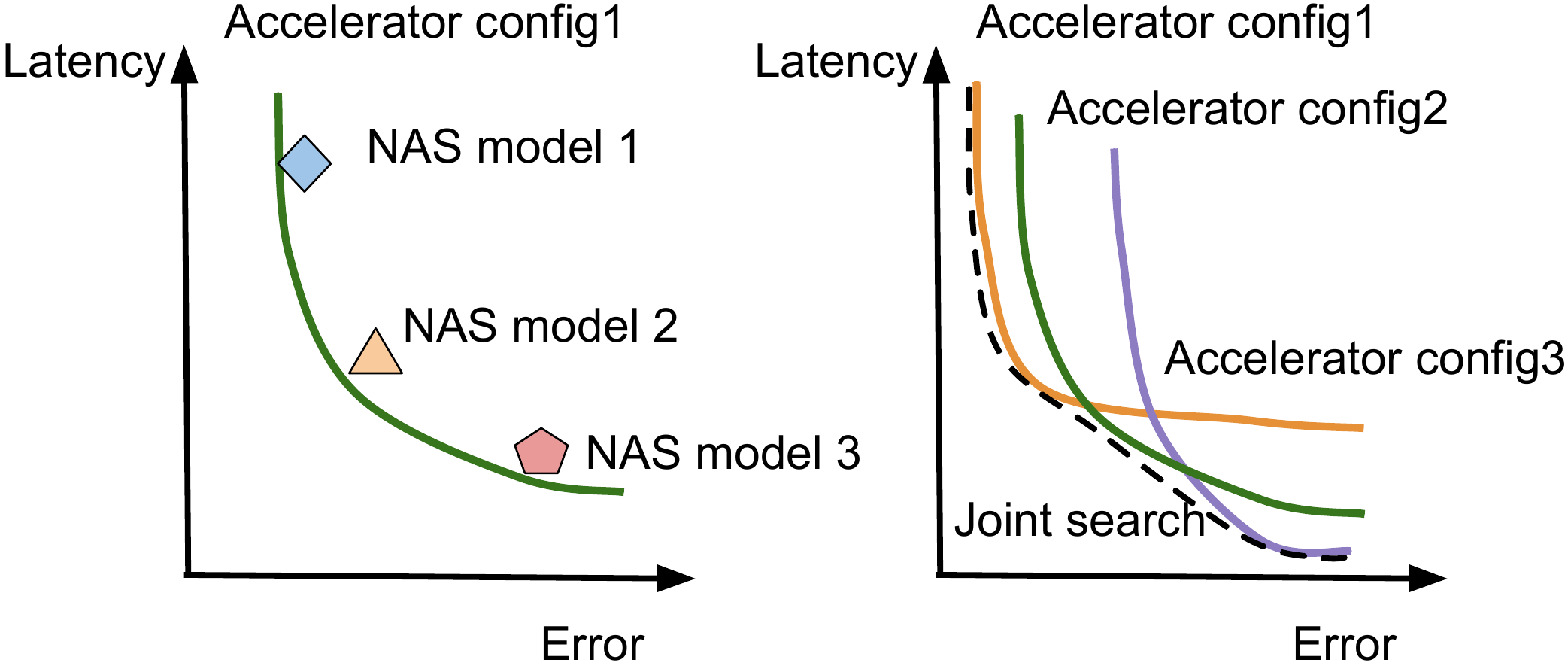}
 \caption{Different accelerator configurations have different Pareto frontiers consisting of different NAS models. Joint search effectively extends the Pareto frontier by joining multiple frontiers.}
    \label{fig:motivation}
\end{figure}


Figure~\ref{fig:motivation} shows that while conventional platform-aware NAS selects models along the Pareto frontier with different latency and accuracy tradeoffs for one target device, NAHAS further expands the Pareto frontier by enabling different hardware accelerator configurations a high-level workflow of NAHAS.
To summarize our contributions:
\begin{itemize}
    \item We develop a fully automated framework that can jointly optimize neural architectures and hardware accelerators. For the first time, we demonstrate the effectiveness of co-optimizing neural architecture search with the parameterization of a highly optimized industry-standard edge accelerator.

    \item We propose a latency-driven search method, which is hardware-agnostic and achieves state-of-the-art results across multiple search spaces. NAHAS outperforms platform-aware search including MnasNet~\citep{MnasNet2018} and EfficientNet~\citep{EfficientNet2019}, and manually crafted models including MobileNet-v2-1.4 and Manual-EdgeTPU~\citep{xiong2020mobiledets,AcceleratorAwareNAS2020}, on all latency targets by around 1\% in ImageNet top-1 accuracy or around 20\% in inference latency.

    \item We observe that different NAS search spaces consisting of different fundamental operations, coupled with accuracy and inference latency objectives, require completely different accelerator configurations. Co-adapting NAS search spaces and accelerator configurations for different model sizes and latency targets becomes essential for domains such as autonomous driving, edge computing, etc.

    \item We compare different optimization strategies and find that joint search consistently outperforms optimizing NAS and HAS in an alternating fashion or in a nested loop. Oneshot search can significantly reduce search cost (total number of samples and search time); however, it is less suitable for large models when constructing the super network is too costly.

\end{itemize}

\section{Related Work}\label{sec:related_work}

\textbf{ML-driven Hardware Accelerator Search:}
Hierarchical-PABO~\citep{bo:frontiers:2020} and FlexiBO~\citep{flexibo:arxiv:2020} leveraged multi-objective Bayesian optimization for neural network accelerator design.
In order to reduce computational costs, \citep{cnn_gen:cyber:2020} applied a genetic algorithm to design CNN models without modifying the underlying architecture.
HyperMapper~\citep{prac_dse:mascots:2019} used random forests in automatic tuning of hardware accelerator configuration in a multi-objective setting.

\textbf{Platform-Aware \& Efficient NAS:} There has been growing interest in leveraging NAS to automate the design process of edge models~\citep{mnv3:iccv:2019,MnasNet2018,ChamNet2019}.
ProxylessNAS directly searches for ImageNet by proposing a gradient-based approach to train binarized parameters~\citep{proxylessNAS2019}.
FBNet proposed a differentiable platform-aware NAS using Gumbel Softmax~\citep{FBNet2019}. NetAdapt~\citep{yang2018netadapt} and AMC~\citep{he2018amc} utilize a direct metric like latency to finetune the number of channels of a pre-trained model. Accelerator-aware NAS~\citep{AcceleratorAwareNAS2020} targets industry-standard accelerators and identified state-of-the-art (SoTA) models for an edge TPU. However, none of these works optimize the underlying hardware accelerators together with NAS.
There are many recent efforts to reduce the search cost of NAS~\citep{brock2018smash,bender2018oneshot,pham2018efficient,liu2019darts,cai2020onceforall,dong2019search,proxylessNAS2019,FBNet2019}.
Existing works are heavily focusing on highly regularized search spaces consisting of mainly Inverted Bottleneck (IBN) layers, for FLOPs on mobile CPUs.
Less attention has been paid to the adaptation of search space or a latency/power driven search for modern accelerators such as DSPs or EdgeTPUs.

\textbf{Hardware Metrics:} Platform-aware NAS typically searches for efficient models with lower parameter counts and FLOPs (number of multiply-add operations). However, optimizing \emph{indirect metrics} like parameter counts or FLOPs won't necessarily improve the \emph{direct metrics} of latency~\cite{bender2020can,FBNet2019}.\looseness-1. For example, a multi-branch network like NASNet~\citep{NASNet2017} has lower parameter counts and FLOPs compared to the layer-wise network such as ResNet and MobileNet~\citep{sandler2018mobilenetv2}, but its fragmented cell-level structure is not hardware friendly and is often slow on real-world devices and accelerators. The effects of \emph{indirect metrics} like parameter counts and FLOPs do not have a simple correlation with \emph{direct metrics} such as latency and power. For example, the model can run out of memory if the number of parameters is too large.

\textbf{Co-design:}
There is a growing body of work exploring neural architecture search and hardware design~\citep{jiang2020standing,choi2020dance,jiang2020hardware,achararit2020apnas,coexploration2020,kwon2018co,yang2019synetgy}. However, most of the work targets FPGAs or academic accelerators~\cite{Eyeriss2016}, where the hardware is less optimized on real workloads (e.g. ImageNet) and the performance model is less accurate. NAHAS, however, targets a highly optimized industry-standard ML accelerator and the performance simulator is a validated, cycle-accurate model. Moreover, instead of using a fixed NAS search space, NAHAS demonstrates effectiveness on a real ImageNet workload, on multiple search spaces including an adapted search space for EdgeTPU. NAHAS is the first to adapt the NAS search space for the co-design of NAS and accelerators.

\section{Methodology}\label{sec:method}

In this section, we will generalize the ideas of joint optimization of hardware and model architectures with NAHAS and present two different approaches. 

\subsection{Formulation}
The objective for NAHAS is to find a neural architecture configuration $\alpha$ and hardware accelerator configuration $h$ such that the validation accuracy can be maximized while meeting a chip area and latency target.

\begin{align}
 & \min\limits_{\alpha, h} \gL (\alpha, h, \omega_{\alpha}^{*}, \sD_{val}) \\
\mathrm{s.t.} ~& ~ \omega_{\alpha}^{*}=\arg\min_{\omega_{\alpha}} \gL(\alpha, h, \omega_{\alpha}, \sD_{train}),\\
\mathrm{and} ~& ~Latency(\alpha, h) \leq T_{latency}, ~Area(h) \leq T_{area},
\end{align}

where $\gL$ indicates the objective function of the tasks (e.g., cross-entropy for classification) and $\omega_{\alpha}$ denotes the weights of the architecture $\alpha$.
NAHAS introduces a broader search space than NAS (neural architecture search) or HAS (hardware accelerator search) alone, with the flexibility to fix either $\alpha$ or $h$ therefore, the optimization problem is reduced to NAS or HAS.
We empirically compare different optimization strategies in Section~\ref{subsec:compared-sota}.

\subsection{NAS Search Space}

\subsubsection{IBN-based Search Space}\label{subsubsec:ibn_ss}

\textbf{MobileNetV2:} We build the architecture search space $\gS_{1}$ based on the standard MobileNetV2. The search space is tailored for mobile edge processors, and therefore consists mostly of efficient operations such as mobile IBN. Specifically, we search for the kernel size from \{3, 5, 7\} for each IBN layer, and we also search for the expansion ratio from \{3, 6\} for each block except for the first one, which has the default expansion ratio of 1.
In MobileNetV2, there are 17 inverted residual blocks, and thus the cardinality of $\gS_{1}$ is about 8.4e12.

\textbf{EfficientNet:} In order to create larger NAS models and to better leverage modern edge accelerators which have larger number of compute units and memory capacities, we build the architecture search space $\gS_{2}$ based on the standard EfficientNet-B0. Similar to $\gS_{1}$, we also search for the kernel size from \{3, 5, 7\} and the expansion ratio from \{3, 6\}. Since there are 16 inverted residual blocks in EfficientNet-B0, the cardinality of $\gS_{2}$ is about 1.4e12.

\subsubsection{Evolved NAS Search Space}\label{subsec:evolved_ss}

\begin{figure}[!t]
  \centering
\includegraphics[width=0.95\linewidth]{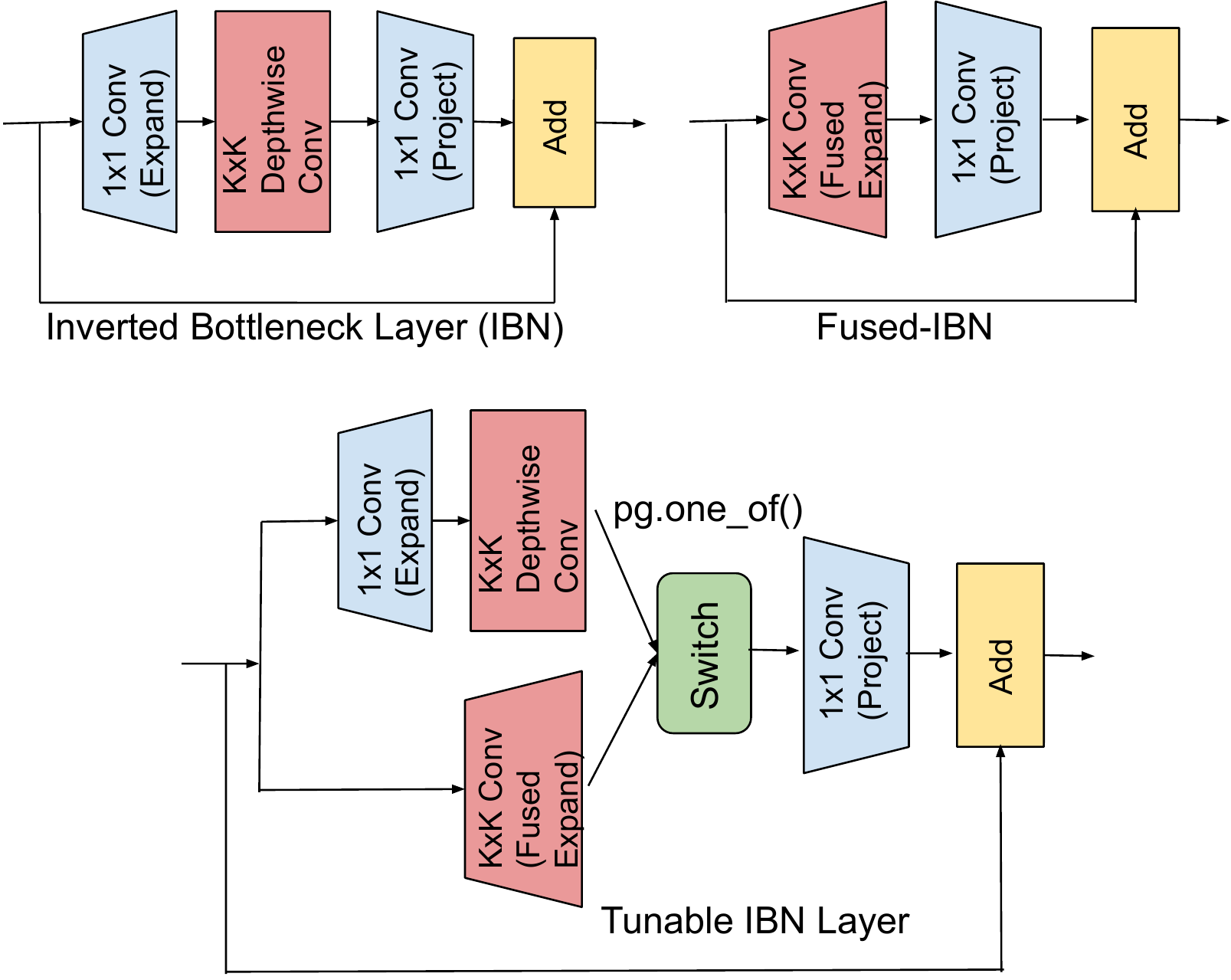}
    \caption{Tunable Layers for Edge Accelerators. In a Fused-IBN layer, we use a $one\_of$ operation in PyGlove to select whether to use a regular depthwise convolution or a fused convolution.}
\label{fig:edgetpu-ops}
\end{figure}

IBNs remain the predominant building blocks in many SoTA mobile models.
The previously described search spaces, built around MobileneV2 and EfficientNet in Section~\ref{subsubsec:ibn_ss}, are both IBN based.
Although IBN layers are good at reducing the parameter count and FLOPs, depthwise-separable convolutions in IBN can significantly increase inference latency or reduce accelerator utilization.
For example, many industry standard edge accelerators, such as Google's EdgeTPU and Qualcomm DSPs, are designed for regular convolutions with sufficient compute intensity (more computations per data transfer)~\citep{zhou2018resourceefficient}.
For certain tensor shapes and kernel dimensions, a regular convolution can utilize the hardware up to 3x more efficiently than the depth-wise variation on an EdgeTPU despite the much larger computation cost (7x more FLOPs)~\citep{xiong2020mobiledets}. This leads to our decision to re-engineer the search space for edge accelerators.

\textbf{Fused IBN Layers:} We adopted the fused IBN layer from MobileDets~\citep{xiong2020mobiledets} that modifies an IBN layer by fusing together the first $1$x$1$ convolution (which usually comes with an expansion ratio) and its subsequent $K$x$K$ depthwise convolution with a single $K$x$K$ full convolution.
We keep the expansion factor as conventional MobileNet models to enable channel size expansion. We name this \textit{Fused-IBN}.
Fused-IBN provides more trainable parameters over baseline IBN but runs faster on edge accelerators for some tensor shapes.

\begin{figure}[!t]
  \centering
\includegraphics[width=\linewidth]{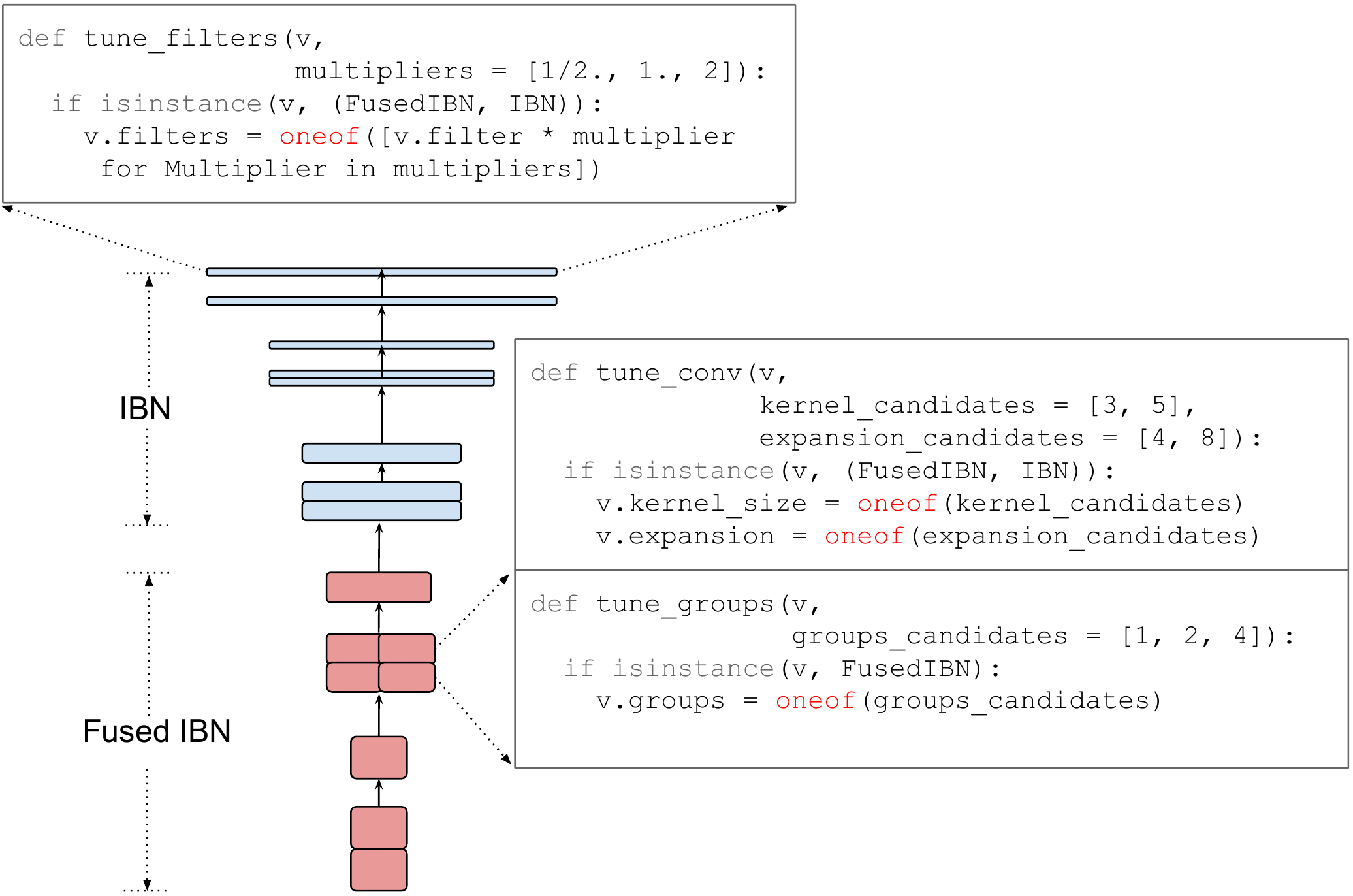}
    \caption{Evolved Search Space: NAHAS respects Efficientnet's compound scaling ratios, while introducing edge efficient Fused-IBN with tunable parameters. We use a symbolic programming library named PyGlove~\cite{peng2021pyglove} to tune \textbf{filter size, kernel size, expansion factor, and groups}.}
\label{fig:search_space}
\end{figure}

\textbf{Tunable Layers via Symbolic Programming:} Intuitively, Fused-IBN becomes less efficient when the network gets deeper and the channel dimension gets larger, due to the cost of full convolution over large input channels. We introduce a more flexible, tunable search space via PyGlove~\cite{peng2021pyglove} symbolic library, to select ops type at each layer as well as selecting layer parameters such as kernel size, expansion factor, filter scaling multiplier, and number of groups. This flexibility provides a middle ground between IBN and Fused-IBN layers.\looseness=-1

Figure~\ref{fig:edgetpu-ops} demonstrates idea of a switched Fused-IBN layer.
For some tensor shapes, Fused-IBN can be more effective when running on edge accelerators, such as early layers with smaller channel sizes in a convolutional network.
However, as the network goes deeper and channel size becomes larger, we may still adopt the conventional IBN to reduce latency and cost.
Figure~\ref{fig:search_space} demonstrates our evolved search space, which consists of a mixture of Fused-IBN, and conventional IBN layers. In our manually crafted model (\textit{Manual-EdgeTPU} presented in our evaluation results) based on the evolved search space, we use a fixed number of fused-IBN in the early layers.
However, in NAHAS, we use PyGlove~\cite{peng2021pyglove} to symbolize each layer as a tunable search space. For example, we can replace any static node in a computational graph, with a tunable node with a predefined search space (e.g., filter scaling, number of groups, kernel size, and expansion factor, etc.).
Therefore, we can essentially transform any static neural network into a tunable search space.
In the following sections, we evaluate NAHAS on multiple search spaces, leveraging the powerful symbolic programming.

\subsection{Accelerator Search Space}\label{subsec:has_search_space}

\begin{figure}[th!]
  \centering
\includegraphics[width=\linewidth]{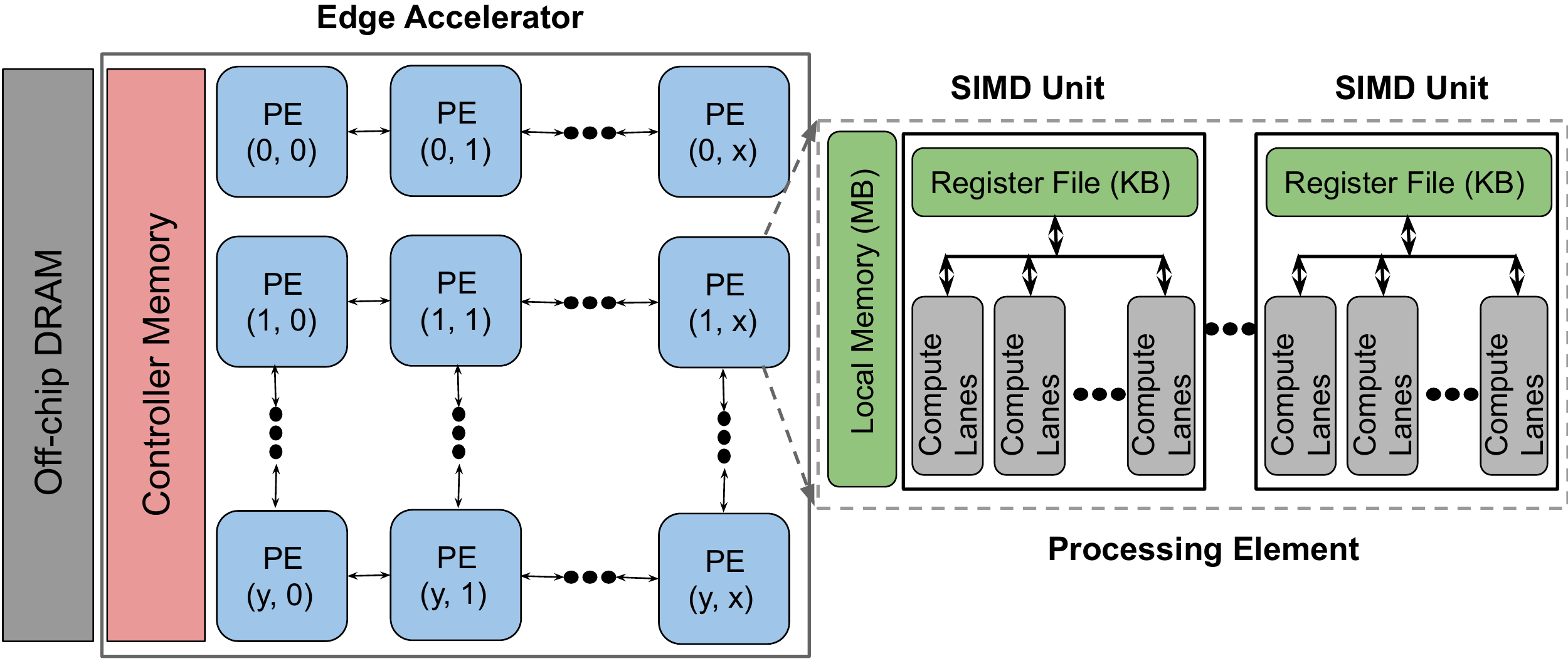}
    \caption{Edge Accelerator Configuration Details.}
\label{fig:edge_accelerator}
\end{figure}

The target device is an industry-standard, highly parameterized edge accelerator which allows us to create various configurations in a large design space with tradeoffs between performance, power, area, and cost. The accelerator in Figure~\ref{fig:edge_accelerator} features a set of parallel processing elements (PE) organized in a 2D tile. The number of PEs in each dimension determines the aspect ratio of the chip. In each PE there are multiple compute lanes that shares a local memory and each lane has a register file and a series of single-instruction multiple-data (SIMD) style multiply-accumulate (MAC) compute units. Each of these architecture components provide a degree of parallelism and a corresponding area cost.
With a fixed chip area budget, the hardware accelerator search (HAS) optimizes the on-chip resource allocation, balancing the compute and memory and searching for the best chip parameterization for a given application.

\begin{table}[h!]
  \begin{center}
  \setlength{\tabcolsep}{3.5pt}
    \label{tab:table1}
    \begin{tabular}{||l|c|r||} 
    \hline
      \textbf{parameters} & \textbf{type} & \textbf{search space}\\
      \specialrule{.15em}{.05em}{.05em}
      PEs\_in\_x\_dimension & int & 1, 2, 4, 6, 8 \\
      PEs\_in\_y\_dimension & int & 1, 2, 4, 6, 8 \\
      SIMD\_units & int & 16, 32, 64, 128 \\
      compute\_lanes & int & 1, 2, 4, 8 \\
      local\_memory\_MB & int & 0.5, 1, 2, 3, 4 \\
      register\_file\_KB & int & 8, 16, 32, 64, 128 \\
      io\_bandwidth\_gbps & float & 5, 10, 15, 20, 25 \\\hline
    \end{tabular}
    \caption{Edge Accelerator Search Space.}
  \end{center}
\end{table}

As our baseline we take the default accelerator configuration which is optimized while considering a series of production workloads from multiple domains. The baseline configuration features 4x4 PEs where each PE has 2 MB local memory and 4 compute lanes. Each compute lane has a 32 KB register file and 64 4-way SIMD units. Since the accelerator is targeted for edge use cases, SIMD units can sustain the peak throughput for 8-bit quantized operations. This baseline configuration can deliver a peak throughput of 26 TOPS/s at 0.8 GHz.

Unlike the NAS search space, the HAS search space contains many invalid points, which makes training a cost model or joint search with the in-house simulator more challenging. Invalid configurations can be caused by many reasons. For example, the created accelerator configuration in combination with the NAS model may not be supported by the compiler or a NAS model may be created that is too large for the generated HAS configuration, etc.

\subsection{Search Objective}\label{subsec:search-objective}
Chip power and area are two important efficiency metrics for accelerator design. In this work, we focus on maximizing model accuracy while meeting an inference latency constraint on a target device. The latency constraint can be easily swapped with an energy (average power times latency) constraint that considers both power and latency. We also impose a chip area constraint, which is set equal to the baseline accelerator design. This paper does not focus on minimizing chip area or latency beyond these set constraints. In a more general setting, the chip resource constraints can be set to area, power, energy, or a combination of many, however, experimentation on the mixture of constraints is not the focus of this paper.

Similar to MnasNet~\citep{MnasNet2018}, we use a customized weighted product to encourage Pareto optimal solutions. More specifically, we have a \textit{optimization metric} of model accuracy and two \textit{hardware constrain metrics} of model latency and chip area. The optimization goal is to maximize model accuracy while meeting the latency and area constraints.
\begin{align}
    \max\limits_{\alpha, h} Accuracy(\alpha, h)   (\frac{Latency(\alpha, h)}{T_{latency}})^{w_{0}}  (\frac{Area(h)}{T_{area}})^{w_{1}},
\end{align}

where $w_{0}, w_{1}$ are the weight factors:
\begin{align}
w_{0} &=
\begin{cases}
p,  & \textrm{if} ~ Latency(\alpha, h) \leq T_{latency} \\
q,  & \textrm{otherwise}
\end{cases},\\
w_{1} & =
\begin{cases}
p, & \textrm{if} ~ Area(h) \leq T_{area} \\
q, & \textrm{otherwise}
\end{cases}
\end{align}
When $p=0, q = -1$, the reward function imposes a hard latency constraint and we simply use accuracy as the objective if the measured latency is meeting the latency target $T$ and only sharply penalize the objective value if the sample violates the latency constraint. When $p=q=-0.07$\footnote{\cite{MnasNet2018} found -0.07 to empirically ensure Pareto-optimal solutions have similar reward under different accuracy-latency
trade-offs.}, the reward becomes a soft constraint function. In the NAHAS evaluations, we use both rewards for different experiments. 

\subsection{Search Strategy}\label{subsec:search_strategy}
\subsubsection{Joint search w/o weight sharing}
Similarly as NASNet~\citep{NASNet2017} and MnasNet~\citep{MnasNet2018}, we use PPO~\citep{ppo} as the controller algorithm to optimize the joint search space from NAS and HAS. The controller samples the search space using a recurrent network, each sample is trained by a child program. For the MobileNetV2 search space, we use a proxy task that trains each sample on ImageNet for only 5 epochs and it takes 5000 samples for the controller to converge. For larger models using EfficientNet search space, training the proxy task for 15 epochs while reducing the total number of samples to 2000 improves the results. 

\subsubsection{Joint search with weight sharing}
\label{subsec:oneshot}

To further reduce the search cost, we employ an efficient search method with weight sharing. Similarly as ProxylessNAS~\citep{proxylessNAS2019} and TuNAS~\citep{bender2020can}, we use the controller decisions from the NAS space to construct a super-network for optimizing the architecture, meanwhile using the decisions from the HAS space to create a sub-graph for computing the cost. Decision points from both spaces are optimized by a RL algorithm within the same graph. For each training step, we train the model weights and the controller decision points in an interleaved way. To achieve better results, we apply the absolute reward function and RL warm-up procedure introduced in TuNAS. 

To estimate the latency of the model on a given accelerator configuration, we train a cost model with random generated samples using an in-house accelerator simulator. We need a cost model because as NAS becomes much faster with oneshot search, the query to the accelerator performance simulator for chip area and inference latency becomes the new bottleneck for NAHAS oneshot search. Given a sampled neural architecture configuration and an accelerator configuration, the cost model predicts the accelerator area $f_{a}(h)$ and model accuracy $f_{l}(\alpha, h)$. We use an MLP network with ReLU to encapsulate the non-linearity in the latency prediction.
The area predictor and latency predictor largely share parameters with only separate parameterization in the prediction heads.
\begin{align}
Loss = MSE(L_{a}, f_{a}(h)) + \lambda MSE(L_{l}, f_{l}(\alpha, h)),
\end{align}
The cost model was trained with 500k labeled data randomly generated by permuting the neural architecture configurations and accelerator configurations. Because labeled data for accelerator performance is much cheaper than labeled data for NAS accuracy, and the collection of data can utilize the vast amount of CPU resources, we do not consider the cost of training a cost model. We use a 3-layer MLP of hidden size 256 and apply a dropout of 0.1 to mitigate overfitting at each layer.



\section{Evaluation}

\subsection{Experiment Setup}
\textbf{Accelerator Performance Simulator:} Evaluating the candidate neural and hardware architectures accurately is a key requirement for the NAHAS framework.
Simply using the number of MACs or parameters of the neural model as a proxy for performance can be highly inaccurate as its performance highly depends on how the neural network is mapped on the hardware architecture and its unique compute characteristics. To this end, we have utilized an in-house cycle-accurate performance simulator and an analytical area model based on hardware synthesis. We deployed both of these estimators as a service where multiple NAHAS clients can send parallel requests. This provides a flexible way to scale-up the performance and area evaluations.

\textbf{Search Hyperparameters:}
For the end2end search, we choose PPO as it is tested by time. We use the average performance of 10 trials as a reward.
We use Adam optimizer with the learning rate of 0.0005 to update the controller, where the policy gradients are clipped by 1.0.
For each trial, we train the sampled candidate by five epochs using RMSProp.
For these five epochs, we will first warm up the model by two epochs using the learning rate from 0 to 0.66, and then cosine decay it from 0.66 to 0 for the rest three epochs.
For the oneshot search, we utilize REINFORCE to optimize the controller following TuNAS. We use Adam with a learning rate of 0.0048 to optimize it and use the momentum as 0.95 for baseline.
In addition, we use RMSProp to optimize the shared weights following the same learning rate schedule as TuNAS.
The latency predictor is pre-trained.

\textbf{Cost Model Hyperparameters:} The cost model was trained with hyperparameters defined in Table~\ref{tab:cost-model}.
In the oneshot search, we replace the accelerator performance simulator with the trained cost model and search for the best model for five different latency targets (0.3ms, 0.5ms, 0.8ms, 1.1ms, and 1.3ms). The average error between the latency target and the estimated latency of the best model using the accelerator performance simulator is only 0.4\%. 

\begin{minipage}{\textwidth}
  \begin{minipage}[b]{0.49\textwidth}
    \centering
    \includegraphics[width=0.9\textwidth]{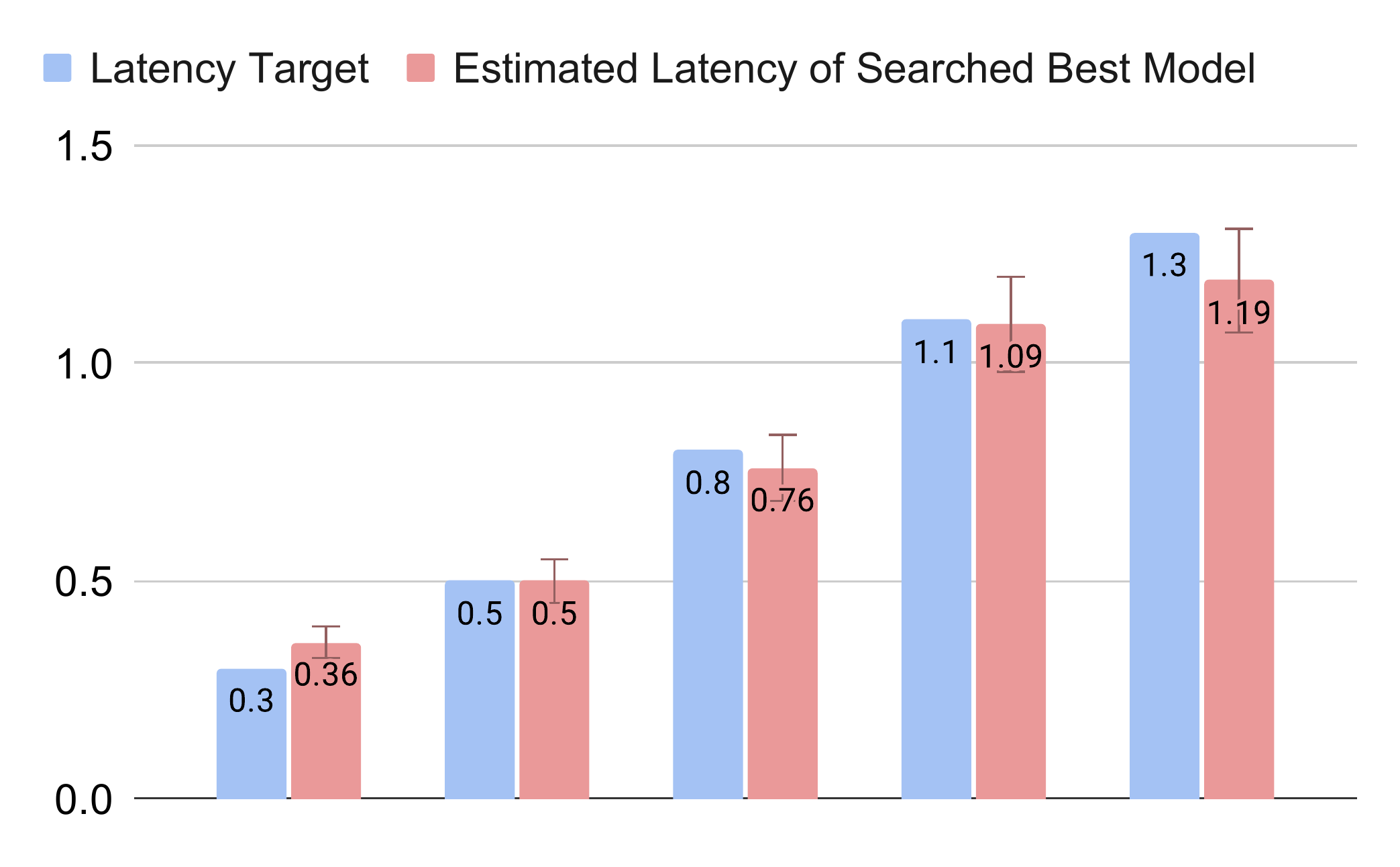}
    \captionof{figure}{Cost Model Accuracy.}
  \end{minipage} \\
  \hfill
  \begin{minipage}[b]{0.49\textwidth}
    \centering
    \begin{tabular}{||l|c|r|l||} 
    \specialrule{.15em}{.05em}{.05em}
      Optimizer & Adam \\
      Loss Re-weight $\lambda$ & 10 \\
      Batch size & 128 \\
      Hidden dimension & 256 \\
      Learning rate & 0.001 \\
      Training steps & 600k \\
      Input feature size & 394 \\
      \specialrule{.15em}{.05em}{.05em}
    \end{tabular}
      \captionof{table}{
      Cost Model Hyperparameters.
      }
      \label{tab:cost-model}
    \end{minipage}
\end{minipage}

\subsection{Sample Distributions}
\begin{figure}[hbt!]
  \centering 
    \label{subfig:nahas-platformaware-efficientnetb0}
    \includegraphics[width=0.9\linewidth]{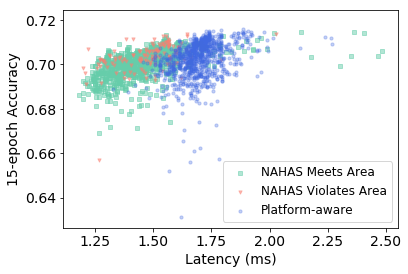}
    \caption{
    Comparison of the sample distributions during search.
    }
    \label{fig:sample-distribution}
\end{figure}

To study the sample distribution during search, we first compare our NAHAS search with previous platform-aware NAS, as shown in Figure~\ref{fig:sample-distribution}. Search is performed on  EfficientNet-B0 with Squeeze-and-Excite layers and Swish non-linearity. In platform-aware NAS, the target device is fixed to the baseline accelerator design, as described in Section~\ref{subsec:has_search_space}; whereas in NAHAS search, the target chip area is set to the same as the target device of platform-aware search. Latency targets are set of 1ms. We observe that without the flexibility to change the hardware configuration, platform-aware NAS always converges to sub-optimal solutions of either higher latency or lower accuracy. For NAHAS, not all the samples traversed meet the chip area constraint. However, traversing through samples violating the resource constraints (red points in the figure) can help converge to more pareto-optimal samples eventually with both higher accuracy and lower inference latency.

\subsection{Latency/Energy-Driven NAHAS}

\begin{figure}[hbt!]
  \centering
      \includegraphics[width=0.9\linewidth,trim={0cm 0cm 0cm, 0.2cm},clip]{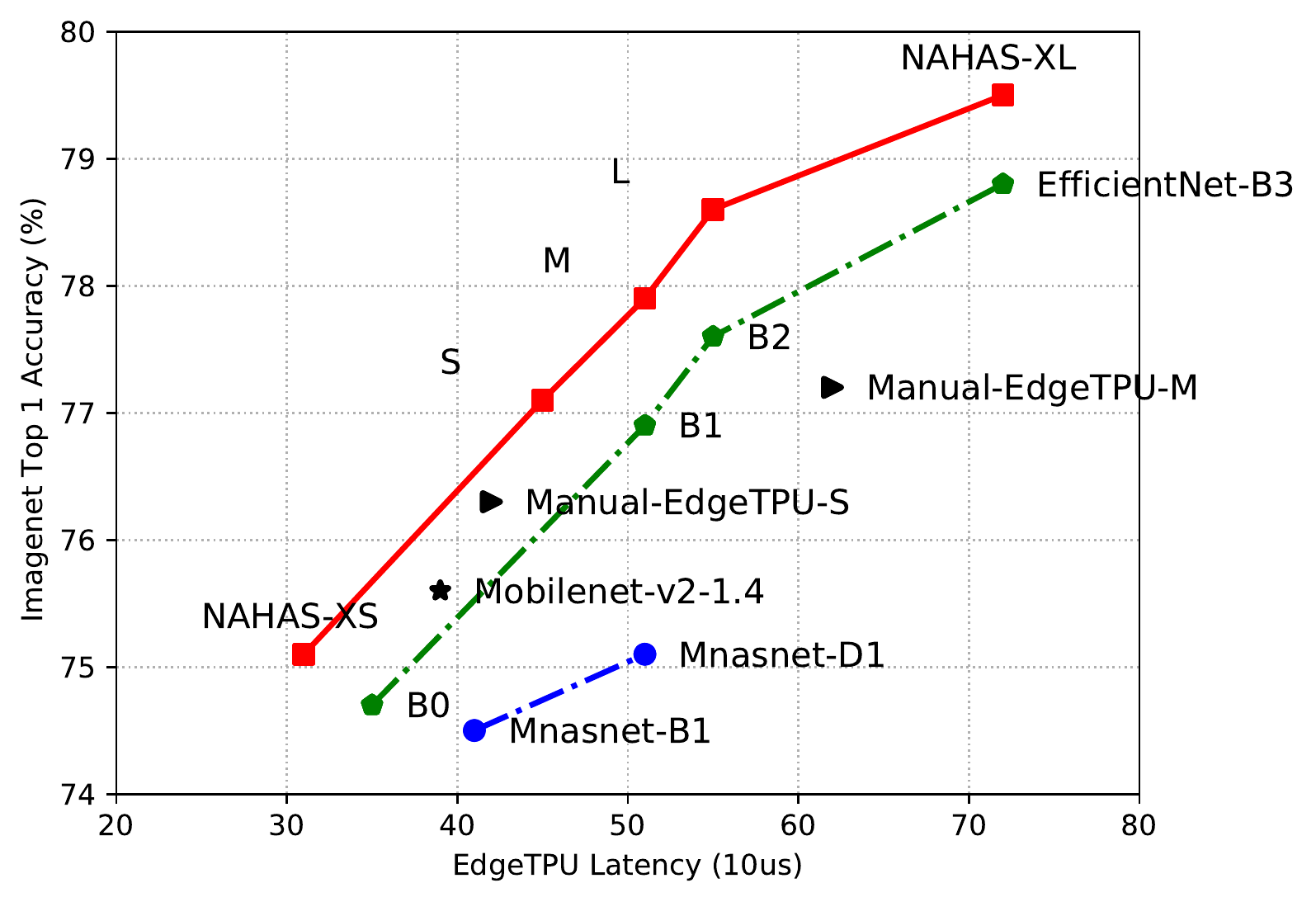}
 \caption{\textbf{Inference Latency vs. ImageNet Accuracy.} NAHAS significantly outperform other platform-aware, efficient ConvNets.}
    \label{fig:latency_nahas}
\end{figure}

Inference latency can be a critical metric for applications with stringent service-level-agreement (SLA) requirements (e.g. mobile face recognition) or real-time requirements (e.g. autonomous driving). In this section, we compare a latency-driven NAHAS based on both the IBN-only search space in Section~\ref{subsubsec:ibn_ss} and adapted search space discussed in Section~\ref{subsec:evolved_ss} with related platform-aware NAS and manually crafted models for EdgeTPUs. Manutal-EdgeTPU-S and Manual-EdgeTPU-M are manually designed models targeting EdgeTPU on the evolved search space described in Section~\ref{subsec:evolved_ss}. However, manually crafted models can be inferior in identifying efficient models with lower inference time, as model latency is a complex function of accelerator resources, compute-to-memory ratio, and neural model characteristics (operation types and tensor shapes). Without an active interaction with an accurate accelerator performance simulator, it is extremely hard to predict model latency given an arbitrary neural model architecture and a target accelerator configuration.

Demonstrated in Figure~\ref{fig:latency_nahas}, \textbf{NAHAS consistently outperforms related work by around 1\% ImageNet top-1 accuracy at all latency targets. With the same accuracy, NAHAS on average reduces inference latency by around 20\%}. We empirically found that a IBN-only search space is good for identifying small, low-latency models (e.g. NAHAS-XS, NAHAS-S) while the proposed evolved search space is good for identifying larger, more accurate models (e.g. NAHAS-M, NAHAS-L, NAHAS-XL) with more relaxed latency targets. Manually crafted models can be very efficient at a particular latency target (e.g. Manual-EdgeTPU-S) however, they can be less efficient when comparing the entire latency spectrum (e.g. Manual-EdgeTPU-M), which manifests the importance of automation.

Apart from inference latency, accelerator energy consumption is another important factor when co-optimizing neural architectures and accelerators. Mobile processors have a stringent power budget due to limited battery life, therefore, maximizing model accuracy while meeting a power budget is another useful way to co-design neural architectures and accelerators. We evaluated a energy-driven NAHAS, where energy is a product of power and latency, comparing to related mobile-efficient NAS and Manual-EdgeTPU. We found that \textbf{our method can improve energy efficiency by up to 2x comparing to related work at the same ImageNet top-1 accuracy}, as shown in Figure~\ref{fig:first_page_results}.

\begin{table*}[!ht]
  \begin{center}
    \scalebox{0.85}{
    \begin{tabular}{l|c|c|c} 
    \specialrule{.15em}{.05em}{.05em} 
      \multirow{2}{*}{Model} & \multirow{2}{*}{Top-1 Acc.} & \multirow{2}{*}{Latency in \textbf{ms} (Ratio-to-best)} & \multirow{2}{*}{Energy in \textbf{mJ} (Ratio-to-best)}\\
      & & &  \\
      \toprule
      EfficientNet-B0~\citep{EfficientNet2019} wo SE/Swish & 74.7\% & 0.35 (1.17x) & 1.00 (1.64x) \\
      MobileNetV2~\citep{sandler2018mobilenetv2} & 74.4\% & 0.30 (1.00x) & 0.70 (1.15x) \\
      MnasNet-B1~\citep{MnasNet2018} & 74.5\% & 0.41 (1.37x) & 0.88 (1.44x) \\
      ProxylessNAS~\citep{proxylessNAS2019} & 74.8\% & 0.42 (1.40x) & 0.98 (1.61x) \\
      Manual-EdgeTPU-small & 76.2\% & 0.42 (1.40x) & 1.78 (2.91x) \\
      IBN-only fixed accelerator & 74.6\% & 0.38 (1.27x) & 0.82 (1.34x) \\
      IBN-only NAHAS multi-trial & 74.9\% & \textbf{0.30} & 0.75 (1.23x)\\
      IBN-only NAHAS oneshot & \textbf{76.5\%} & 0.35 (1.17x) & \textbf{0.61} \\
      \midrule \midrule
      EfficientNet-B1~\citep{EfficientNet2019} wo SE/Swish & 76.9\% & 0.51 (1.04x) & 1.50 (1.53x) \\
      MnasNet-D1~\citep{MnasNet2018} & 75.1\% & 0.55 (1.12x) & 1.75 (1.78x)\\
      Fixed accelerator multi-trial w fused-IBN & 75.3\% & 0.52 (1.06x) & 1.32 (1.35x) \\
      IBN-only NAHAS multi-trial & 77.4\% & 0.52 (1.06x) & 1.50 (1.53x) \\
      NAHAS multi-trial w fused-IBN & \textbf{77.9\%} & 0.51 (1.04x) & 1.12 (1.14x) \\
      IBN-only NANAS oneshot & 76.8\% & \textbf{0.49} & \textbf{0.98} \\
      \midrule \midrule
      EfficientNet-B3~\citep{EfficientNet2019} wo SE/Swish & 78.8\% & 0.72 (1.12x) & 2.28 (1.56x) \\
      Manual-EdgeTPU-medium & 77.2\% & \textbf{0.62} & 2.72 (1.86x) \\
      MobilenetV3 w SE & 76.8\% & 1.44 (2.32x) & 4.00 (2.74x) \\
      Fixed accelerator multi-trial w fused-IBN & 78.2\% & 0.74 (1.19x) & 1.75 (1.20x) \\
      NAHAS multi-trial w fused-IBN & \textbf{79.5\%} & 0.72 (1.12x) & \textbf{1.46} \\
    \bottomrule 
    \end{tabular}
    }
  \caption{
  Comparison on accuracy v.s. latency with previous approaches. Models are grouped into different regimes and sorted by accuracy.
  On the fast latency regime, ``NAHAS oneshot'' has the best performance, while on the high accuracy regime, ``NAHAS multi-trial'' achieves the best performance.
  }
  \label{tab:relatedwork}
  \end{center}
\end{table*}
\begin{table*}[!htb]
  \begin{center}
    \label{tab:compare-citiscape}
    \scalebox{0.85}{
    \begin{tabular}{l|c|c|c} 
    \specialrule{.15em}{.05em}{.05em} 
      \multirow{2}{*}{Model} & \multirow{2}{*}{mIOU Acc.} & \multirow{2}{*}{Latency in \textbf{ms} (Ratio-to-best)} & \multirow{2}{*}{Energy in \textbf{mj} (Ratio-to-best)}\\
      & & &  \\
      \toprule
      EfficientNet-B0 wo SE/Swish & 73.8\% & 3.29 (1.08x) & 6.71 (1.08x) \\
      EfficientNet-B1 wo SE/Swish & 72.8\% & 3.66 (1.20x) & 8.04 (1.29x) \\
      EfficientNet-B2 wo SE/Swish & 72.6\% & 3.71 (1.21x) & 8.48 (1.36x) \\
      Manual-EdgeTPU-S & 71.2\% & 4.15 (1.36x) & 13.31 (2.14x) \\
      Manual-EdgeTPU-M & 74.4\% & 4.75 (1.55x) & 17.68 (2.84x)\\
      IBN-only NAHAS multi-trial & 73.6\% & \textbf{3.06} & 6.54 (1.05x) \\
      NAHAS multi-trial w fused-IBN & \textbf{74.8}\% & 4.17 (1.36x) & \textbf{6.23} \\

    \bottomrule
    \end{tabular}
    }
  \caption{
  Comparison on Cityscapes Segmentation Dataset~\cite{Cordts2016Cityscapes}.
  }
  \end{center}
\end{table*}

\subsection{Detailed Comparison with SoTA}\label{subsec:compared-sota}

We compare NAHAS with mobile-efficient models (including NAS-based and manually crafted) and NAHAS variants on fixed accelerator NAHAS (NAS only with default accelerator configuration), multi-trial NAHAS (without parameter sharing) and oneshot NAHAS (with parameter sharing). We categorize the models into three: small, medium, and large. We use a latency target of 0.3ms, 0.5ms, and 0.7ms and use a energy target of 0.7mJ, 1.0mJ, and 1.5mJ for these three categories. According to Table~\ref{tab:relatedwork}, NAHAS significantly outperforms related work on all three perspectives (accuracy, latency, and energy), except for latency when comparing to medium sized manually crafted Manual-EdgeTPU. However, Manual-EdgeTPU is much worse (1.82x) in energy than NAHAS. We conclude that:
\setlist{nolistsep}
\begin{itemize}[noitemsep]
\itemsep0em 
    \item Adapting hardware accelerator configurations helps identify more pareto efficient models in terms of model accuracy, inference latency, and energy consumption, which cannot be identified by a fixed hardware NAS.
    \item The NAS search space also needs to be adapted to create more edge efficient neural architectures. For example, removing SE and Swish non linearity significantly improves inference latency and energy while not impairing accuracy significantly. Some expensive ops such as fused-IBN, can be more effective when running on edge accelerators, for certain tensor shapes.
    \item Different neural architectures with different performance targets lead to drastically different accelerator configurations. For example, larger models require a higher memory-to-compute ratio in the accelerator design. NAHAS identifies edge accelerator configurations with larger number of processing elements (PE) and smaller memory capacity, compared to the baseline edge accelerator, for small models with very tight latency/energy target; It identifies accelerator configurations with larger local memory and register file size for large models and models containing fused-IBN. 
    \item Model inference time and energy consumption on the target accelerator is a complex function of accelerator resources, chip area, neural architectures, tensor shapes, that is intricate to capture with manual engineering. The manually designed Manual-EdgeTPU can be competitive on one hardware performance metric, but can be significantly worse in the other. 
    \item Oneshot search is more effective than a multi-trial search without parameter sharing for smaller models with lower than 0.5ms latency. However, it has limited performance when the model becomes larger with a more relaxed latency/energy target. Constructing a super-network in a oneshot search can be impractically too expensive when the search space is larger or the model size becomes larger. 
    
\end{itemize}

\vspace{-0.5em}
\subsection{Ablation Study}\label{subsec:ablation}

\begin{figure}[hbt!]
  \centering 
    \label{subfig:joint-vs-phase}
      \includegraphics[width=0.85\linewidth]{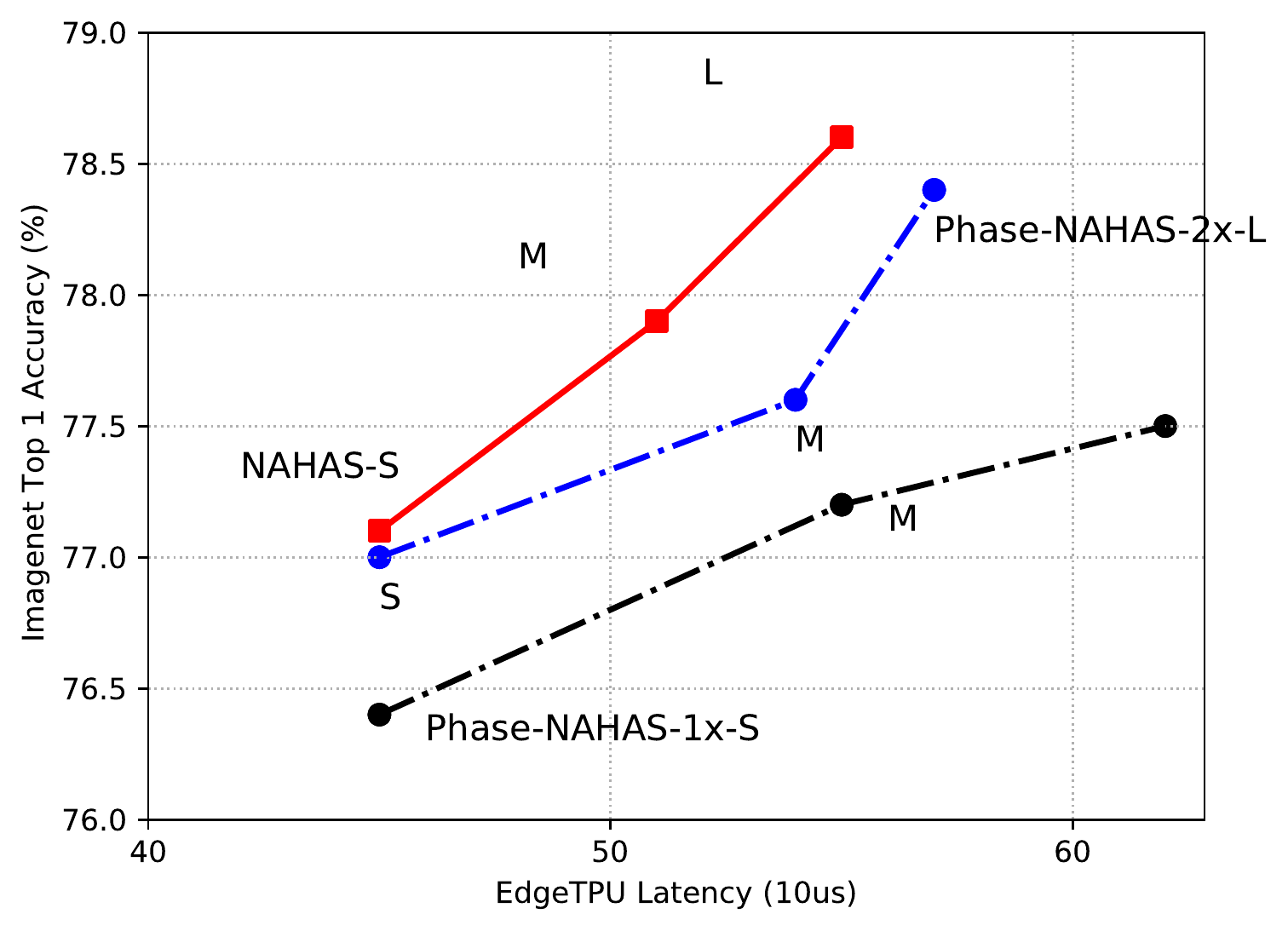}
    \caption{Joint Search vs. Phase-based Search. 
    }
    \label{fig:search_method} 
\end{figure}

\textbf{Search Method:} Figure~\ref{fig:search_method} compares joint search with phase search. More particularly, in a phase-based NAHAS, we start with a hardware accelerator search on a fixed initial neural architecture in the search space (MobileNetV2, EfficientNet-B1, and EfficientNet-B2) with a soft constraint function described in Section~\ref{subsec:search-objective}, aiming to find an accelerator configuration which is pareto-optimal in terms of latency and chip area. Then we apply a NAS with a hard constraint function described in Section~\ref{subsec:search-objective} on the selected best accelerator configuration, aiming to identify good neural architectures that strictly meet the hardware latency constraint.
We compare Phase-NAHAS using 1x and 2x total searched samples, compared to the NAHAS multi-trial baseline.
\vskip -0.15in

NAHAS phase search with the same number of searched samples performs much worse than the NAHAS multi-trial joint search. Doubling the total searched samples thus the search time, improves the quality of results. However, we find that the initial neural architecture creates a large variance in search quality.
      
\textbf{Segmentation Results:}\label{sec:segmentation} We also conduct extensive experiments on segmentation task to verify the generalization ability of our proposed NAHAS. Specifically, we use the popular CityScapes dataset~\cite{Cordts2016Cityscapes} to benchmark model performance. We train all models from scratch with a SGD optimizer and a cosine learning rate of 0.08 and warmup proportion of 0.01. We use a training batch size of 64 and evaluation batch size of 32. During the NAHAS search, we use a proxy training of 20 epochs. In the final evaluation, we train the models for 1000 epochs.

\section{Conclusion}\label{sec:conclusion}

We systematically studied the importance and strategies of jointly optimizing neural architectures and hardware accelerators. Our identified NAHAS models outperform related work by 1\% in ImageNet top-1 accuracy and reduce accelerator energy consumption by up to 2x. 

\bibliography{example_paper}
\bibliographystyle{icml2021}





\end{document}